\documentclass[runningheads]{llncs}

\usepackage{eccv}

\usepackage{eccvabbrv}

\usepackage{graphicx}
\usepackage{booktabs}

\usepackage[accsupp]{axessibility}  

\usepackage{booktabs} 
\usepackage{graphicx} 
\usepackage{amsmath}  
\usepackage{amssymb}  
\usepackage{pifont}   
\usepackage{xcolor}   
\usepackage{caption}
\usepackage{colortbl}  
\usepackage{wrapfig}
\usepackage{cuted}

\definecolor{bestbg}{RGB}{198,234,212}   
\definecolor{secondbg}{RGB}{231,242,255} 
\definecolor{secondbg2}{RGB}{255, 218, 224}
\definecolor{mygreen}{HTML}{00AA00}

\newcommand{\second}[1]{\cellcolor{secondbg}#1}

\newcommand{\boldparagraph}[1]{\noindent\textbf{#1}\ }

\usepackage{hyperref}

\usepackage{orcidlink}

\begin{document}







\title{ProLaViT: Learning Progressive Latent Visual Thoughts in Structured Latent Space}

\titlerunning{ProLaViT: Progressive Latent Visual Thoughts}

\author{Peiming Li\inst{1}\textsuperscript{*} \and
Yifan Wang\inst{1}\textsuperscript{*} \and
Xiaotian Zhang\inst{2} \and
Zhiyuan Hu\inst{3} \and
Shiyu Li\inst{1} \and \\
Zheng Wei\inst{1}\textsuperscript{\dag} \and
Yang Tang\inst{1}\textsuperscript{\ddag\dag}}

\authorrunning{P.~Li et al.}

\institute{Basic Algorithm Center, PCG, Tencent\\
\email{\{peimingli, wyattyfwang, shyuli, hemingwei, ethanntang\}@tencent.com} \and
Zhejiang University\\
\email{xiaotian.24@intl.zju.edu.cn} \and
Peking University\\
\email{zhiyuanhu@stu.pku.edu.cn}}

\maketitle
\let\thefootnote\relax\footnotetext{\textsuperscript{*}Equal contribution. \quad \textsuperscript{\dag}Corresponding author. \quad \textsuperscript{\ddag}Project Lead.}

\begin{abstract}
Multimodal Large Language Models (MLLMs) have achieved remarkable progress but still struggle with complex visual reasoning tasks requiring multi-step perception and logical deduction. 
While explicit visual generation incurs prohibitive computational costs, existing latent approaches often rely on external experts or lack rigorous cognitive logic. 
In this paper, we introduce ProLaViT (\textbf{Pro}gressive \textbf{La}tent \textbf{Vi}sual \textbf{T}hought), a framework empowering MLLMs to perform structured visual derivation in the continuous latent space. 
Unlike works dependent on heterogeneous external models, ProLaViT leverages an endogenous self-distillation mechanism, utilizing the model's own visual encoder to supervise latent thoughts. 
To facilitate this, we construct a scalable programmatic synthesis pipeline enabling the model to internalize algorithmic precision without inference-time tools. 
We design two reasoning paradigms: (1) Coarse-to-Fine Causal Chain for spatial tasks, guiding attention from global context to local targets. (2) Dialectical Reasoning Chain for logical tasks, incorporating counterfactual thinking for verification. 
Furthermore, we propose a Distance-Weighted Diversity Loss to impose topology-aware constraints, preventing feature degeneration by enforcing semantic distinctiveness. 
Extensive experiments demonstrate that ProLaViT outperforms baselines on vision-centric benchmarks, achieving superior accuracy and interpretability with high efficiency.
\end{abstract}

\begin{figure*}[t]
    \centering
    \includegraphics[width=0.9\textwidth]{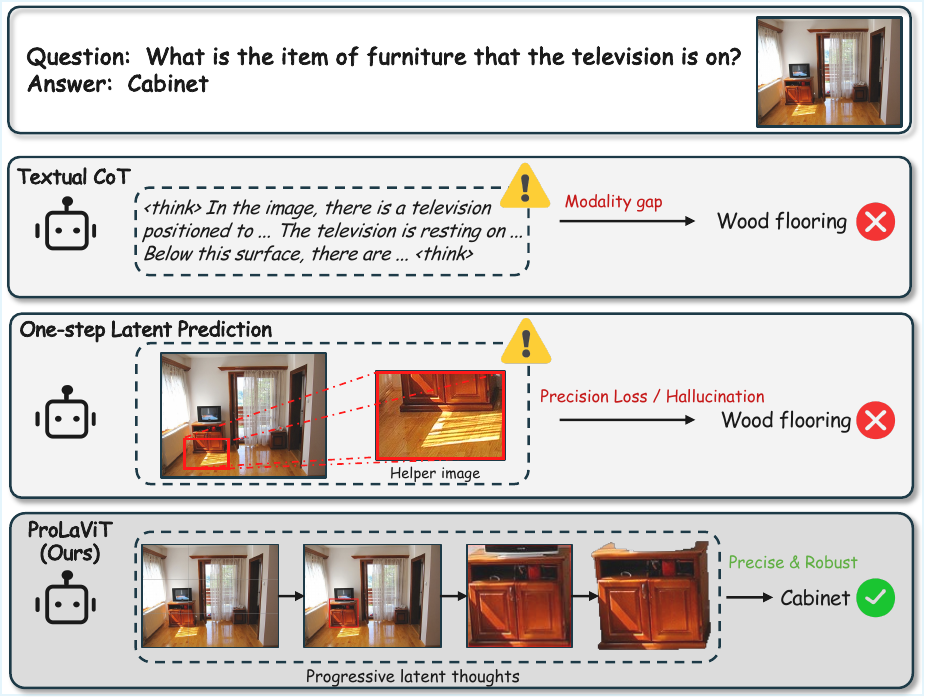} 
    \caption{\textbf{Comparison of multimodal reasoning paradigms.} 
    \textbf{Top:} Textual CoT suffers from a \textit{modality gap}, failing to accurately ground spatial relationships (e.g., mistaking the wood flooring for the cabinet). 
    \textbf{Middle:} One-step Latent Prediction attempts to identify the target region in a single leap but suffers from \textit{precision loss}, often attending to irrelevant background areas (hallucination). 
    \textbf{Bottom:} \textbf{ProLaViT (Ours)} decomposes the task into a \textit{progressive latent chain} (Locate $\to$ Focus $\to$ Isolate), enabling precise target isolation and robust reasoning.}
    \label{fig:intro}
\end{figure*}

\section{Introduction}
\label{sec:intro}

The advent of Multimodal Large Language Models (MLLMs)~\cite{liu2024visual, bai2023qwen, achiam2023gpt} has revolutionized the intersection of vision and language, enabling systems to describe images and answer open-ended questions with unprecedented fluency. 
Nevertheless, despite these achievements, current MLLMs frequently exhibit limitations when addressing visual reasoning tasks that necessitate multi-step spatial perception, geometric analysis, or logical deduction~\cite{tong2024eyes, fu2024blink, cvbench, chartqa}. 
The core challenge lies in the reasoning mechanism, specifically in how to effectively bridge the gap between high-level linguistic semantics and low-level visual features.

Prevalent approaches rely on Textual Chain-of-Thought (CoT)~\cite{wei2022chain, zhang2023multimodal} to decompose complex tasks. 
However, as illustrated in Fig.~\ref{fig:intro}, reasoning solely in text suffers from a fundamental modality gap. 
Text is inherently abstract and often fails to capture fine-grained spatial nuances. 
For instance, when asked to identify the furniture beneath a television, the model fails to ground the spatial relationship, hallucinating ``wood flooring'' instead of the correct ``cabinet''. To address this, recent works have explored incorporating visual information into the reasoning chain. Explicit visual generation methods such as Anole~\cite{anole}, Bagel~\cite{bagel}, Zebra-CoT~\cite{zebra-cot} and ThinkMorph~\cite{thinkmorph2024} generate intermediate pixel-level images to aid reasoning. While effective, the computational cost of iterative image generation is prohibitive for real-time applications. Conversely, implicit latent reasoning approaches~\cite{li2025latent, yang2025machine, hao2024training, qin2024chain, ilvr, monet} attempt to model reasoning within the continuous feature space. However, as illustrated in Fig.~\ref{fig:intro}, these approaches typically adhere to a One-step Latent Prediction paradigm. Attempting to directly localize a specific visual region such as a helper image crop within a single step often exceeds the representational capacity of the model. This limitation frequently results in significant precision loss where the model focuses on irrelevant background elements like the floor rather than the intended target, ultimately leading to fragile representations and erroneous predictions.

In this work, we argue that effective visual reasoning requires neither expensive pixel generation nor unstructured latent guessing, but rather \textbf{Progressive Visual Derivation} within a structured latent space. We introduce \textbf{ProLaViT}, a novel framework that decomposes complex visual tasks into a structured chain of latent thoughts. Instead of jumping to a conclusion, it mimics human cognitive processes, following a \textit{Locate $\to$ Focus $\to$ Isolate} causal chain to progressively refine visual attention, ensuring precise object isolation and robust reasoning.

A key challenge in training such multi-step latent models is the absence of ground truth for intermediate images. Unlike prior works that rely on external vision experts such as~\cite{yang2024depth, kirillov2023segment, oquab2023dinov2}, ProLaViT employs an \textbf{Endogenous Self-Distillation} mechanism. This approach effectively avoids the domain gaps and pipeline complexity often introduced by external dependencies. We leverage the MLLM's own frozen vision encoder to extract features from a sequence of auxiliary images. Crucially, these auxiliary sequences are generated via a scalable programmatic synthesis pipeline. By training on these rigorous, code-generated visual trajectories, ProLaViT effectively internalizes algorithmic precision, learning to perform latent operations in its latent space without needing external tools during inference.

Furthermore, a common pathology in sequential latent reasoning is \textit{latent collapse}, where the representations of subsequent steps become indistinguishable, degrading the reasoning chain into redundancy. 
To mitigate this, we propose a novel \textbf{Distance-Weighted Diversity Loss}. 
This objective function explicitly constrains the topology of the latent space, penalizing cosine similarity between reasoning steps. 
Crucially, the penalty is weighted by the causal distance in the chain, forcing the latent state of a segmentation step to be significantly distinct from the initial global view step, thereby ensuring that each step contributes unique and progressive information to the inference process. Our main contributions are summarized as follows:
\begin{itemize}
    \item We propose ProLaViT, a framework that enables MLLMs to perform Progressive Visual Derivation in the latent space, overcoming the efficiency bottlenecks of explicit generation and the precision limitations of unstructured latent methods.

    \item We introduce an Endogenous Self-Distillation strategy supported by a programmatic data synthesis pipeline, allowing the model to internalize rigorous algorithmic logic without external vision experts.

    \item We design a Distance-Weighted Diversity Loss to prevent latent collapse, enforcing semantic distinctiveness between reasoning steps proportional to their causal distance.

    \item Extensive experiments demonstrate that ProLaViT achieves SOTA performance, offering a superior balance of accuracy, interpretability, and efficiency.
\end{itemize}

\section{Related Work}

\subsection{Multimodal Chain-of-Thought Reasoning}
Chain-of-Thought (CoT) prompting~\cite{wei2022chain} enhances LLM reasoning by decomposing problems into logical steps. Multimodal CoT~\cite{zhang2023multimodal, hao2025can, jiang2025mme} extends this to vision-language tasks, and works like Qwen-VL~\cite{bai2023qwen} and LLaVA-CoT~\cite{xu2024llavacot} demonstrate that reasoning data improves performance. However, these methods rely on projecting visual information into discrete text. Recent studies~\cite{fu2024blink, tong2024eyes} indicate this creates a modality gap, as abstract text often fails to capture fine-grained spatial nuances. This leads to hallucinations where reasoning is linguistically coherent yet visually inaccurate. In contrast, ProLaViT performs reasoning directly in the continuous latent space, preserving visual fidelity.

\subsection{Visual-Augmented Reasoning with External Tools}
To bridge the modality gap, researchers have augmented MLLMs with external tools~\cite{wu2023visual, shen2024hugginggpt, suris2023vipergpt, yang2023mmreact, lu2023chameleon} or explicit visual generation~\cite{thinkmorph2024, anole, dong2024dreamllm, ge2024seed, koh2024generating}. Specifically, methods like CoVT~\cite{qin2024chain, shao2024visual} employ experts such as SAM~\cite{kirillov2023segment} or DepthAnything~\cite{yang2024depth} for supervision. However, these approaches face two critical limitations: (1) \textit{Prohibitive Cost} due to pixel-level generation or external inference latency; and (2) \textit{Pipeline Complexity} arising from heterogeneous dependencies. In contrast, ProLaViT adopts an \textit{endogenous self-distillation} mechanism, leveraging the MLLM's own frozen encoder to supervise latent thoughts, eliminating external overhead.

\subsection{Latent Visual Reasoning}
Recent works explore reasoning within continuous latent spaces to balance efficiency and expressiveness. While language-centric methods~\cite{hao2024training, codi} introduce continuous tokens, they lack spatial grounding. Visual adaptations like Mirage~\cite{yang2025machine} and LVR~\cite{li2025latent} map visual semantics into LLMs but typically adopt an \textit{unstructured} or \textit{single-step} paradigm. As our analysis reveals, this often leads to \textit{latent collapse}, where intermediate representations degenerate into redundancy. ProLaViT advances this direction by introducing Structured Progressive Derivation. Instead of black-box transitions, we enforce a rigorous causal chain (\textit{Locate $\to$ Focus $\to$ Isolate}) constrained by a Distance-Weighted Diversity Loss, ensuring that latent thoughts remain topologically distinct and logically progressive.

\section{Method}
\subsection{Overview}
\begin{figure*}[ht]
    \centering
    \includegraphics[width=\textwidth]{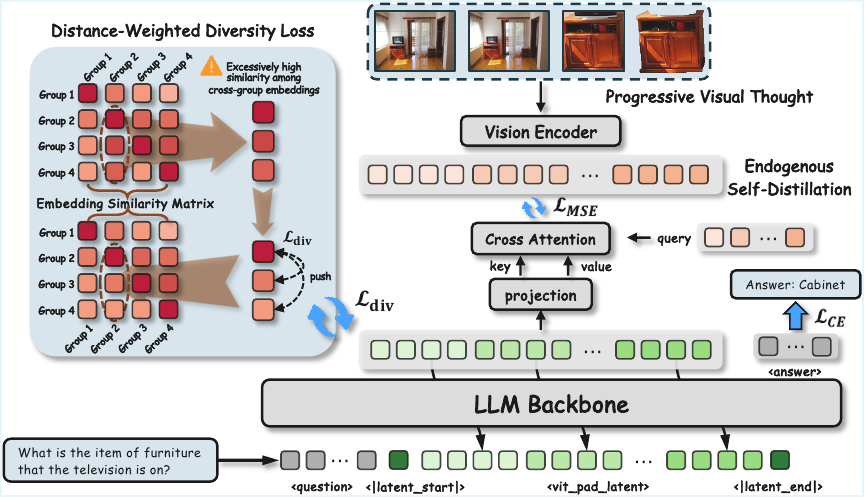} 
    \caption{\textbf{Overview of ProLaViT.} The framework generates progressive latent thoughts (green tokens) supervised by an Endogenous Self-Distillation mechanism (\textit{Right}), where latent states are aligned with visual features from synthesized auxiliary images via $\mathcal{L}_{MSE}$. To prevent representation degeneration, a Distance-Weighted Diversity Loss ($\mathcal{L}_{div}$, \textit{Left}) penalizes excessive similarity between reasoning steps, enforcing topological distinctiveness in the latent space.}
    \label{fig:architecture}
\end{figure*}
\begin{figure*}[t]
    \centering
    \includegraphics[width=0.95\textwidth]{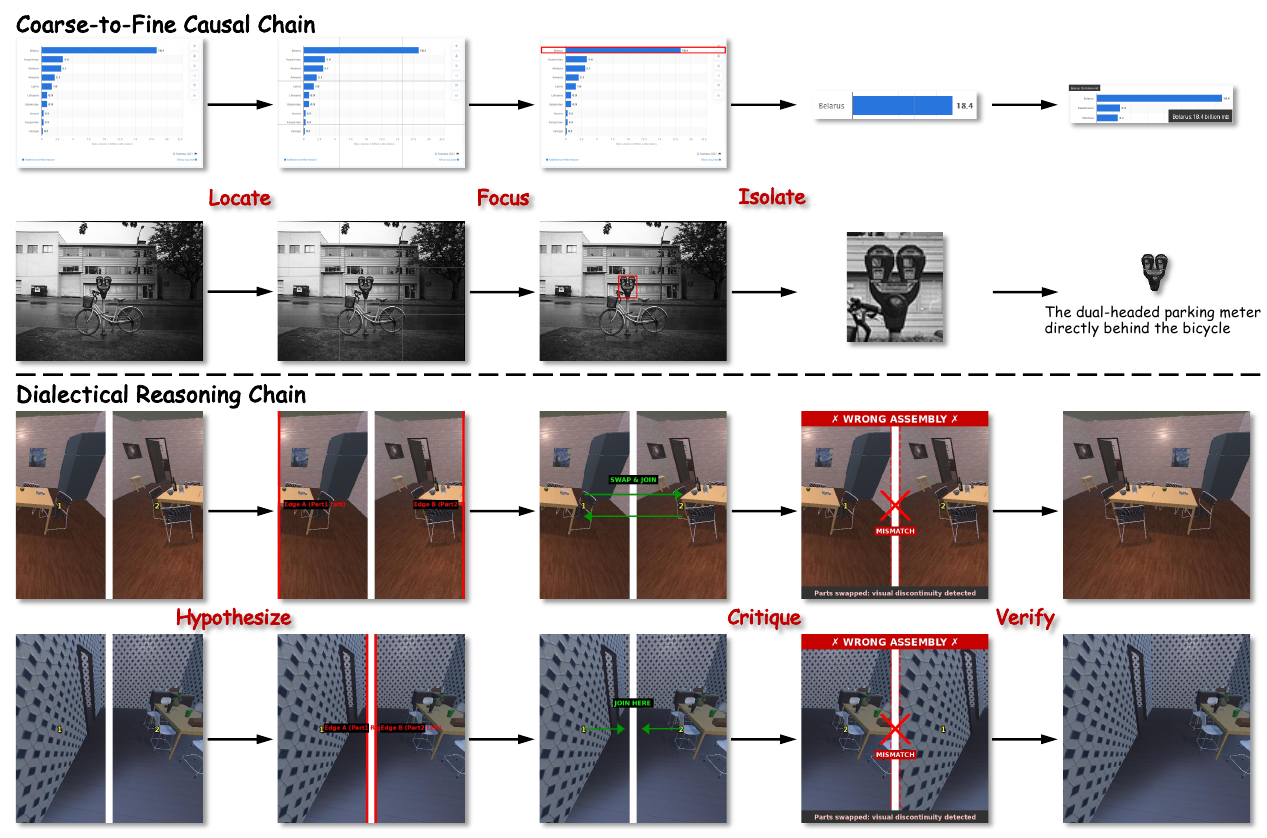} 
    \caption{\textbf{Illustration of the two reasoning paradigms in ProLaViT.} \textbf{Top:} The \textit{Coarse-to-Fine Causal Chain} for spatial tasks progressively narrows attention from global context to specific targets (\textit{Locate $\to$ Focus $\to$ Isolate}). \textbf{Bottom:} The \textit{Dialectical Reasoning Chain} for logical tasks (e.g., jigsaw puzzles) employs a trial-and-error approach (\textit{Hypothesize $\to$ Critique $\to$ Verify}) to validate visual consistency.}
    \label{fig:reasoning_chains}
\end{figure*}
Fig.~\ref{fig:architecture} illustrates the overall architecture of ProLaViT. Given an input image $I$ and a textual query $Q$, ProLaViT performs visual reasoning through a progressive latent thought chain rather than a single-step prediction. The framework consists of three core components: (1) A \textbf{Progressive Latent Visual Thought} module that decomposes complex visual tasks into a structured sequence of $K$ intermediate reasoning steps. (2) An \textbf{Endogenous Self-Distillation} mechanism that leverages the model's own vision encoder to provide supervision signals. (3) A \textbf{Distance-Weighted Diversity Loss} that prevents latent collapse by enforcing topology-aware constraints on the reasoning chain. Formally, let $\mathcal{M} = (\mathcal{V}, \mathcal{L})$ denote a Multimodal Large Language Model with vision encoder $\mathcal{V}$ and language model $\mathcal{L}$. Given an input image $I$ and query $Q$, ProLaViT generates a sequence of latent thoughts $\mathcal{Z} = \{z_1, z_2, \ldots, z_K\}$ that progressively refine the visual understanding before producing the final answer $A$.

\subsection{Progressive Latent Visual Thought}
Unlike prior methods that attempt to predict complex visual cues in a single step, ProLaViT decomposes visual reasoning into a structured chain of latent thoughts, as illustrated in Fig.~\ref{fig:reasoning_chains}. We design two distinct reasoning paradigms tailored for different task types.

\noindent\textbf{Coarse-to-Fine Causal Chain.} For spatial tasks such as visual search and object localization, we adopt a \textit{Locate $\rightarrow$ Focus $\rightarrow$ Isolate} paradigm (Fig.~\ref{fig:reasoning_chains}, Top). Each reasoning step progressively narrows the attention scope:
\begin{equation}
    z_k = \mathcal{T}_k(z_{k-1}, I, c_k), \quad k \in \{1, 2, \ldots, K\},
\end{equation}
where $\mathcal{T}_k$ denotes the $k$-th reasoning transformation, and $c_k$ represents the task-specific context. Specifically, we implement four canonical operations that instantiate this paradigm:
\begin{itemize}
    \item \textbf{Grid View ($z_1$):} Establishes global context with spatial grid overlay (Preparation).
    \item \textbf{Bounding Box ($z_2$):} Performs the \textbf{Locate} step via region localization.
    \item \textbf{Crop ($z_3$):} Executes the \textbf{Focus} step for detailed inspection of the target region.
    \item \textbf{Segmentation ($z_4$):} Completes the \textbf{Isolate} step with fine-grained object boundary delineation.
\end{itemize}

\noindent\textbf{Dialectical Reasoning Chain.} For logical tasks such as jigsaw puzzle assembly, we employ a \textit{Hypothesize $\rightarrow$ Critique $\rightarrow$ Verify} loop that incorporates counterfactual thinking (Fig.~\ref{fig:reasoning_chains}, Bottom). This paradigm enables the model to evaluate multiple hypotheses before committing to a solution. Specifically, we instantiate this loop through four distinct visual operations:
\begin{itemize}
    \item \textbf{Edge Enhancement ($z_1$):} Highlights high-frequency boundary features to identify potential mating surfaces, forming the initial \textbf{Hypothesis}.
    \item \textbf{Directional Guidance ($z_2$):} Visualizes the proposed movement vector using arrow overlays to suggest the assembly orientation.
    \item \textbf{Counterfactual Simulation ($z_3$):} Generates a plausible but incorrect assembly state (e.g., misalignment) to trigger the \textbf{Critique} mechanism, teaching the model to distinguish subtle errors.
    \item \textbf{Solution Verification ($z_4$):} Presents the correctly assembled state to \textbf{Verify} the logical consistency and visual coherence of the final solution.
\end{itemize}

\noindent\textbf{Token Representation.} Each reasoning step is represented by a set of learnable tokens in the LLM's embedding space. Specifically, we allocate $N_{\text{tokens}}$ special tokens (denoted as \texttt{<|vit\_pad\_latent|>}) per reasoning step. For $K=4$ steps with $N_{\text{tokens}}=4$ tokens each, the total latent thought sequence contains $K \times N_{\text{tokens}} = 16$ tokens. These tokens are embedded in the input sequence and trained to capture the intermediate visual states:
\begin{equation}
    \mathbf{H}_{\text{latent}} = \text{LLM}(\mathbf{E}_{\text{input}} \oplus \mathbf{E}_{\text{thought}}),
\end{equation}
where $\mathbf{E}_{\text{thought}} \in \mathbb{R}^{(K \cdot N_{\text{tokens}}) \times d}$ represents the latent thought embeddings, and $\oplus$ denotes concatenation.

\subsection{Endogenous Self-Distillation Mechanism}
A key challenge in training latent reasoning models is the lack of ground truth for intermediate mental images. Prior works rely on external vision experts (e.g., SAM, DINO), which introduces domain gaps and pipeline complexity. Instead, ProLaViT employs an Endogenous Self-Distillation strategy that leverages the MLLM's own frozen vision encoder as the teacher.

\noindent\textbf{Programmatic Synthesis Pipeline.} To generate supervision signals, we construct a scalable programmatic pipeline that produces rigorous visual trajectories. For each training sample, we programmatically generate $K$ auxiliary images $\{I_1^{\text{aux}}, I_2^{\text{aux}}, \ldots, I_K^{\text{aux}}\}$ corresponding to each reasoning step (e.g., applying \texttt{crop(I, bbox)} or \texttt{segment(I, target)}). These auxiliary images are generated automatically using geometric transformations and are guaranteed to be algorithmically precise. Specific implementation details are provided in the Supplementary Materials.

\noindent\textbf{Teacher Feature Extraction.} The frozen vision encoder $\mathcal{V}$ extracts features from each auxiliary image:
\begin{equation}
    \mathbf{F}_k^{\text{teacher}} = \mathcal{V}(I_k^{\text{aux}}), \quad k \in \{1, \ldots, K\},
\end{equation}
where $\mathbf{F}_k^{\text{teacher}} \in \mathbb{R}^{L \times d}$ with $L$ denoting the number of visual tokens per image. Crucially, gradients are not backpropagated through $\mathcal{V}$, ensuring the vision encoder remains frozen during training.

\noindent\textbf{Cross-Attention Knowledge Distillation.} To align the LLM-generated latent thoughts with the teacher features, we employ a single cross-attention over the full sequence. Let $\mathbf{H} \in \mathbb{R}^{N \times d}$ denote the hidden states of all reasoning-step tokens (e.g., $N = K \times n_{\text{tok}}$ with $n_{\text{tok}}$ tokens per step). We project and normalize:
\begin{equation}
    \mathbf{H}^{\text{proj}} = \text{Normalize}\bigl(\text{Linear}(\mathbf{H})\bigr).
\end{equation}
A learnable query matrix $\mathbf{Q} \in \mathbb{R}^{L \times d}$ (one per image length $L$) is expanded to length $K \cdot L$ (e.g., via linear interpolation) so that the number of query positions matches the concatenated teacher sequence. The aligned representation is computed in one cross-attention:
\begin{align}
    \mathbf{\hat{F}} &= \text{CrossAttn}\bigl(\mathbf{Q}_{\text{expand}}, \mathbf{H}^{\text{proj}}, \mathbf{H}^{\text{proj}}\bigr), \quad \mathbf{\hat{F}} \in \mathbb{R}^{K \cdot L \times d},
\end{align}
where $\mathbf{Q}_{\text{expand}} \in \mathbb{R}^{K \cdot L \times d}$. Key and value are both $\mathbf{H}^{\text{proj}}$, thus every query position (across all $K$ steps) attends to the same compressed latent tokens. We split $\mathbf{\hat{F}}$ into $K$ segments of length $L$ and denote the $k$-th segment by $\mathbf{\hat{F}}_k \in \mathbb{R}^{L \times d}$. The self-distillation loss is:
\begin{equation}
    \mathcal{L}_{\text{distill}} = \frac{1}{K} \sum_{k=1}^{K} \bigl\| \mathbf{\hat{F}}_k - \mathbf{F}_k^{\text{teacher}} \bigr\|_2^2.
\end{equation}

\subsection{Distance-Weighted Diversity Loss}
In our experiments, we observe a limitation in the sequential reasoning of ProLaViT, where representations of subsequent reasoning steps gradually become indistinguishable. 
This phenomenon is highlighted by the token similarity matrix in Fig.~\ref{fig:similarity_heatmap}. Such convergence risks degrading the progressive reasoning chain into redundancy, a failure mode we term \textit{latent collapse}.
\noindent To mitigate the aforementioned degradation of the reasoning chain, we propose introducing a diversity loss during training. However, standard diversity losses often prove overly strict, as they indiscriminately penalize any positive similarity. Given that visual features naturally share common patterns (e.g., edges and textures), pursuing complete orthogonality is therefore unrealistic.
Inspired by metric learning, we propose a margin-based diversity loss allowing reasonable similarity while penalizing representation collapse:
\begin{equation}
    \mathcal{L}_{\text{div}}^{\text{margin}} = \frac{1}{|\mathcal{P}|} \sum_{(i,j) \in \mathcal{P}} \max(0, \cos(\mathbf{g}_i, \mathbf{g}_j) - \tau),
\end{equation}
where $\mathbf{g}_k$ denotes the centroid of the $k$-th reasoning step's tokens, $\mathcal{P}$ is the set of all step pairs, and $\tau \in [0, 1]$ is a learnable margin threshold.
A key insight is that reasoning steps with larger causal distance should be more distinct. To capture this, we introduce a distance-weighted penalty:
\begin{equation}
    w_{ij} = \epsilon + (1 - \epsilon) \cdot \left(\frac{d(i, j)}{d_{\max}}\right)^{\alpha},
\end{equation}
where $d(i, j) = |i - j|$ is the causal distance, $d_{\max} = K - 1$, $\epsilon = 0.1$ is the minimum weight, and $\alpha \geq 1$ is a learnable parameter controlling steepness, parameterized as $\alpha = 1 + \text{softplus}(\theta_\alpha)$. The final loss is:
\begin{equation}
    \mathcal{L}_{\text{div}} = \frac{1}{|\mathcal{P}|} \sum_{(i,j) \in \mathcal{P}} w_{ij} \cdot \max(0, \cos(\mathbf{g}_i, \mathbf{g}_j) - \tau).
\end{equation}
This formulation respects the hierarchical structure of the reasoning chain while effectively preventing representation collapse.

\subsection{Training Objective and Strategy}
The overall training objective combines language modeling with latent supervision:
\begin{equation}
    \mathcal{L} = \mathcal{L}_{\text{LM}} + \lambda_{\text{distill}} \mathcal{L}_{\text{distill}} + \lambda_{\text{div}} \mathcal{L}_{\text{div}},
\end{equation}
where $\mathcal{L}_{\text{LM}}$ is the standard cross-entropy loss. We set $\lambda_{\text{distill}} = 1.0$ and $\lambda_{\text{div}} = 0.2$.

To ensure the stable emergence of progressive thoughts and prevent representation collapse, we employ a multi-stage curriculum learning strategy:

\begin{itemize}
    \item \textbf{Phase 1: Latent Anchor Initialization.} 
    We freeze the vision encoder and LLM backbone, training only the embeddings of the special thought tokens. This phase acts as a ``warm-up,'' initializing the latent tokens to a neutral state within the semantic space of the LLM, preventing early optimization instability.
    
    \item \textbf{Phase 2: Endogenous Knowledge Distillation.} 
    We activate the \textbf{Self-Distillation} mechanism ($\mathcal{L}_{\text{distill}}$). Here, the model learns to align its generated latent thoughts with the target visual features extracted by its own frozen encoder. Crucially, we do not yet enforce the diversity constraint, allowing the model to focus solely on accurate feature reconstruction.
    
    \item \textbf{Phase 3: Structured Chain Evolution.} 
    We introduce the \textbf{Distance-Weighted Diversity Loss} ($\mathcal{L}_{\text{div}}$) alongside the distillation loss. This is the critical phase where the reasoning chain transforms from a repetitive sequence into a structured, progressive derivation. The model is forced to differentiate adjacent reasoning steps (e.g., separating \textit{Locate} from \textit{Focus}), establishing the causal topology in the latent space.
    
    \item \textbf{Phase 4: Global Capability Integration.} 
    Finally, we perform joint training with general VQA data. This prevents catastrophic forgetting of general instruction-following abilities while consolidating the learned visual reasoning patterns into the model's global behavior.
\end{itemize}

\section{Experiment}
\subsection{Experimental Setup}
We implement ProLaViT based on Qwen2.5-VL-7B-Instruct. The latent thought uses $K=4$ reasoning steps with $N_{\text{tokens}}=4$ tokens per step. Training is performed on 2 NVIDIA H20 GPUs. For Diversity Loss, $\lambda_{\text{div}}=0.2$, margin $\tau$ initialized to 0.8.

\boldparagraph{Benchmarks.} We evaluate ProLaViT on a comprehensive suite of perception-intensive visual reasoning benchmarks that span multiple dimensions of visual understanding: \textbf{MMVP}~\cite{tong2024eyes}, \textbf{VisPuzzle}~\cite{thinkmorph2024}, \textbf{VStar}~\cite{vstar}, \textbf{ChartQA}~\cite{chartqa}, \textbf{BL\\INK}~\cite{fu2024blink}, \textbf{CV-Bench}~\cite{cvbench}.
\begin{table}[htbp]
\renewcommand{\arraystretch}{1.3} 
\setlength{\tabcolsep}{3.5pt} 
\centering
\large
\caption{Performance comparison on visual reasoning benchmarks. \textbf{ProLaViT (Full)} denotes our model trained with the proposed Distance-Weighted Diversity Loss. All values are reported in accuracy (\%). Best results are in \textbf{bold}, second best are \underline{underlined}.}
\label{tab:main_results}
\resizebox{\textwidth}{!}{
\begin{tabular}{lccccccc|c}
\toprule
\textbf{Method} & \textbf{MMVP} & \textbf{VisPuzzle} & \textbf{VStar} & \textbf{ChartQA} & \textbf{BLINK} & \textbf{CV-Bench} & \textbf{BLINK-J} & \textbf{Avg.} \\
\midrule
\multicolumn{9}{c}{\textbf{\second{\textit{Base Model}}}} \\
Qwen2.5-VL-Instruct & 77.33 & 34.75 & 76.44 & \underline{78.45} & 54.49 & 73.61 & 59.33 & 66.00 \\
\midrule
\multicolumn{9}{c}{\textbf{\second{\textit{Text-Space Reasoning}}}} \\
SFT & 77.00 & 67.50 & 78.01 & 76.74 & 55.60 & 77.36 & 49.33 & 69.86 \\
CoT SFT & 77.66 & 69.75 & 75.91 & 75.83 & 56.54 & 64.55 & 68.66 & 69.18 \\
\midrule
\multicolumn{9}{c}{\textbf{\second{\textit{Latent-Space Reasoning}}}} \\
CoVT & 78.33 & 35.75 & \underline{79.05} & 24.99 & \textbf{57.49} & 75.00 & 66.00 & 61.45 \\
LVR & 77.30 & 35.00 & 76.43 & 64.86 & 53.02 & 76.55 & 57.33 & 64.63 \\
One-step Latent Pred. & \textbf{79.33} & 72.00 & 78.01 & 63.02 & 56.28 & 76.10 & 66.00 & 70.85 \\
\midrule
\multicolumn{9}{c}{\textbf{\second{\textit{Ours (Progressive Latent Visual Thought)}}}} \\
ProLaViT (Base) & 78.00 & 74.00 & \underline{79.05} & 76.18 & \textbf{57.49} & 77.24 & \underline{71.33} & \underline{73.82} \\
ProLaViT + $\mathcal{L}_{\text{div}}^{\text{margin}}$ & 78.33 & \textbf{75.25} & \underline{79.05} & 75.66 & 56.33 & \textbf{78.33} & 67.33 & 73.58 \\
\rowcolor[HTML]{E0E0E0}
ProLaViT + $\mathcal{L}_{\text{div}}$ (Full) & \underline{79.00} & \underline{74.25} & \textbf{80.11} & \textbf{78.99} & \underline{57.23} & \underline{77.66} & \textbf{76.00} & \textbf{75.11} \\
\bottomrule
\end{tabular}
}
\end{table}

\boldparagraph{Baselines.} We compare ProLaViT against the following methods:
\begin{itemize}
    \item \textbf{Qwen2.5-VL-Instruct}~\cite{qwen25vl}: The base MLLM without any fine-tuning, serving as the foundation model.
    \item \textbf{SFT}: Standard supervised fine-tuning on the same training data without latent reasoning.
    \item \textbf{CoT SFT}: Fine-tuning with explicit textual Chain-of-Thought reasoning annotations.
    \item \textbf{CoVT}~\cite{qin2024chain}: Chain-of-Visual-Thought that distills knowledge from external lightweight vision experts.
    \item \textbf{LVR}~\cite{li2025latent}: Latent Visual Reasoning that enables reasoning in the visual embedding space.
    \item \textbf{One-step Latent Prediction}: An ablated variant that predicts all visual features in a single step without progressive decomposition.
\end{itemize}

\subsection{Quantitative Results}
Tab.~\ref{tab:main_results} summarizes the performance comparison across diverse visual reasoning benchmarks. Our proposed framework, ProLaViT (equipped with the full Distance-Weighted Diversity Loss), consistently outperforms baselines, achieving the highest average accuracy of \textbf{75.11\%}.

\boldparagraph{Analysis.} We highlight four key observations from the results:

\begin{enumerate}
    \item \textbf{Structured Decomposition vs. Monolithic Prediction.} 
    Comparing ProLaViT with One-step Latent Prediction, we observe significant gains, particularly on complex tasks like VisPuzzle (+2.25\%) and BLINK-Jigsaw (+10.00\%). This confirms that monolithic latent prediction often hits a performance ceiling due to capacity overload. In contrast, our \textit{Locate $\rightarrow$ Focus $\rightarrow$ Isolate} causal chain allows the model to incrementally refine its attention, leading to more precise visual grounding.
    
    \item \textbf{Bridging the Modality Gap in Spatial Reasoning.} 
    While CoT SFT improves performance on logical puzzles (VisPuzzle: 69.75\%), it suffers a sharp decline on perception-intensive benchmarks (CV-Bench: 64.55\%). This reflects the inherent limitation of text in capturing fine-grained spatial coordinates. ProLaViT overcomes this by reasoning in the continuous latent space, maintaining strong performance across both logical and perceptual tasks without the information loss associated with text projection.
    
    \item \textbf{Robustness via Endogenous Alignment.} 
    A critical failure mode of prior methods is revealed in the ChartQA benchmark. CoVT, which relies on external vision experts (e.g., SAM, DepthAnything), collapses to 24.99\% accuracy, likely due to the domain gap between natural images (on which experts are trained) and abstract charts. ProLaViT avoids this pitfall by employing Endogenous Self-Distillation. By leveraging the MLLM's own vision encoder, our method ensures domain consistency, achieving a robust 78.99\% on ChartQA.
    
    \item \textbf{Efficacy of Dialectical Reasoning.} 
    On the BLINK-Jigsaw benchmark, which demands rigorous logical verification, ProLaViT achieves 76.00\%, a remarkable +16.67\% improvement over the base model. This substantial margin validates the effectiveness of our \textit{Hypothesize $\rightarrow$ Critique $\rightarrow$ Verify} paradigm, demonstrating that structured latent thoughts can successfully simulate trial-and-error processes for complex problem-solving.
\end{enumerate}

\subsection{Qualitative Analysis}
\begin{figure}[ht]
    \centering
    \includegraphics[width=\textwidth]{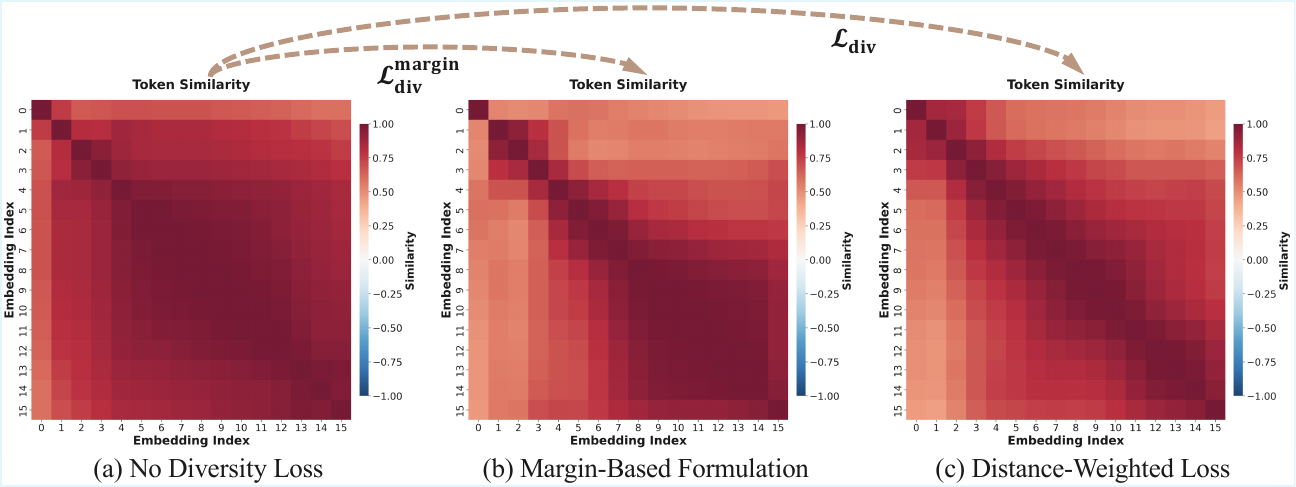} 
    \caption{\textbf{Visualization of latent thought similarity.} We plot the cosine similarity heatmaps between the token representations of different reasoning steps ($z_1$ to $z_4$). \textbf{(a) No Diversity Loss:} The chain suffers from \textit{latent collapse}, where subsequent steps ($z_2$-$z_4$) become indistinguishable. \textbf{(b) Margin-Based Formulation:} Enforces distinctiveness equally across all pairs, potentially disrupting the semantic continuity of adjacent steps. \textbf{(c) Distance-Weighted Loss (Ours):} Respects the hierarchical structure, where adjacent steps retain moderate similarity for context, while distant steps are enforced to be distinct.}
    \label{fig:similarity_heatmap}
\end{figure}

To intuitively understand how ProLaViT structures its reasoning process, we visualize the cosine similarity matrices of the generated latent tokens in Fig.~\ref{fig:similarity_heatmap}. 
\noindent\textbf{Latent Collapse without Constraints.} As shown in Fig.~\ref{fig:similarity_heatmap}(a), in the absence of diversity constraints, the model exhibits severe \textit{latent collapse}. The representations for steps $z_2$ through $z_4$ share excessive similarity, effectively degenerating the progressive chain into redundancy. This explains why the baseline model fails to refine its attention, it is essentially stuck in the initial state.

\noindent \textbf{Margin-Based vs. Distance-Weighted Constraints.} Applying the Margin-Based Formulation (Fig.~\ref{fig:similarity_heatmap}(b)) successfully reduces global similarity, but it indiscriminately penalizes adjacent steps. This disrupts the natural pattern sharing required for causal reasoning. 
In contrast, our Distance-Weighted Diversity Loss (Fig.~\ref{fig:similarity_heatmap}(c)) induces a structure consistent with our hypothesis: adjacent steps maintain moderate similarity to preserve context, while distant steps (e.g., $z_1$ vs. $z_4$) are forced to be distinct. This confirms that ProLaViT learns a non-redundant, progressive derivation path where steps with larger causal distance contribute unique perceptual information.

\subsection{Ablation Study}
To validate the effectiveness of our core design choices, we conduct comprehensive ablation studies focusing on two key aspects: the necessity of progressive decomposition and the impact of the distance-weighted diversity constraint.

\boldparagraph{Effect of Progressive Decomposition.}
We first validate the structural advantage of our multi-step derivation over monolithic prediction. As shown in Tab.~\ref{tab:ablation_structure}, the One-step Latent Prediction baseline, which compresses all visual cues into a single latent state, suffers from a significant performance bottleneck. In contrast, ProLaViT's progressive approach yields substantial gains, particularly on tasks requiring logical depth such as ChartQA (+15.97\%) and BLINK-Jigsaw (+10.00\%). This empirical evidence confirms that complex visual reasoning overloads the capacity of a single latent state, whereas our structured chain (\textit{Locate $\to$ Focus $\to$ Isolate}) effectively distributes the cognitive load, allowing for incremental and precise visual grounding.

\noindent \boldparagraph{Effect of Distance-Weighted Diversity Loss.}
Next, we isolate the impact of our diversity constraints. We first evaluate the Margin-Based Formulation ($\mathcal{L}_{\text{div}}^{\text{margin}}$) without distance weighting. While this formulation mitigates \textit{latent collapse} by penalizing similarity above the threshold $\tau$, it treats all reasoning step pairs equally. As observed in Tab.~\ref{tab:ablation_loss}, this rigid constraint leads to a performance drop on BLINK-Jigsaw (-4.00\%), likely because it disrupts the natural feature sharing (e.g., edges, textures) between adjacent steps.
In contrast, our full Distance-Weighted Diversity Loss ($\mathcal{L}_{\text{div}}$) achieves the best overall performance. By incorporating the causal distance weight $w_{ij}$, it respects the hierarchical structure of the reasoning chain, it assigns lower penalty weights to adjacent steps to preserve context, while enforcing that steps with larger causal distance are more distinct. This design effectively prevents representation collapse while maintaining semantic continuity, as evidenced by the robust improvements on ChartQA (+2.81\%) and BLINK-Jigsaw (+4.67\%).
\begin{table}[ht]
\centering
\small
\setlength{\tabcolsep}{3pt}
\renewcommand{\arraystretch}{0.9}
\caption{Ablation on reasoning structure: \textbf{One-step Prediction} vs. \textbf{Progressive Latent Visual Thought} (Ours). The progressive decomposition yields consistent gains, verifying that complex visual reasoning requires multi-step derivation.}
\label{tab:ablation_structure}
\resizebox{1\textwidth}{!}{%
\begin{tabular}{lcccc|c}
\toprule
\textbf{Structure} & \textbf{VisPuzzle} & \textbf{ChartQA} & \textbf{BLINK-J} & \textbf{CV-Bench} & \textbf{Avg.} \\
\midrule
One-step Latent Pred. & 72.00 & 63.02 & 66.00 & 76.10 & 69.28 \\
\textbf{ProLaViT (Progressive)} & \textbf{74.25} & \textbf{78.99} & \textbf{76.00} & \textbf{77.66} & \textbf{76.73} \\
\midrule
\textit{Improvement} & \textcolor{red!60!black}{+2.25} & \textcolor{red!60!black}{+15.97} & \textcolor{red!60!black}{+10.00} & \textcolor{red!60!black}{+1.56} & \textcolor{red!60!black}{+7.45} \\
\bottomrule
\end{tabular}%
}
\end{table}
\begin{table}[ht]
\centering
\small
\setlength{\tabcolsep}{3pt}
\renewcommand{\arraystretch}{0.9}
\caption{Ablation on diversity constraints. $\mathcal{L}_{\text{div}}^{\text{margin}}$: \textbf{Margin-Based Formulation} (without distance weights). $\mathcal{L}_{\text{div}}$: \textbf{Distance-Weighted Diversity Loss} (Ours).}
\label{tab:ablation_loss}
\resizebox{1\textwidth}{!}{%
\begin{tabular}{lccccc}
\toprule
\textbf{Configuration} & \textbf{VisPuzzle} & \textbf{VStar} & \textbf{ChartQA} & \textbf{BLINK-J} & \textbf{Avg.} \\
\midrule
ProLaViT (No Diversity Loss) & 74.00 & 79.05 & 76.18 & 71.33 & 75.14 \\
\quad + $\mathcal{L}_{\text{div}}^{\text{margin}}$ & 75.25 & 79.05 & 75.66 & 67.33 & 74.32 \\
\rowcolor[HTML]{E0E0E0}
\quad + $\mathcal{L}_{\text{div}}$ (Full) & \textbf{74.25} & \textbf{80.11} & \textbf{78.99} & \textbf{76.00} & \textbf{77.34} \\
\bottomrule
\end{tabular}%
}
\end{table}

\noindent \boldparagraph{Latent Reasoning vs.\ Text-Trace Supervision.}
To disentangle the benefit of latent-space reasoning from the programmatic training data, we convert the same synthesized trajectories into textual traces (\texttt{Crop [x1,y1,x2,y2]$\to$Seg obj}) and fine-tune the base model. Text-Trajectory SFT gains only $+0.9\%$ over CoT SFT (73.41 vs.\ 72.54 avg.), while ProLaViT gains $+4.8\%$ (77.34 avg.). The net $+3.9\%$ arises purely from reasoning in continuous latent space, since text-tokenized coordinates discard sub-pixel appearance cues that ProLaViT's latent chain preserves.

\noindent \boldparagraph{Effect of Endogenous vs. Exogenous Distillation.}
Finally, we examine the necessity of our Endogenous Self-Distillation mechanism. A common alternative is to use external vision models to supervise the latent thoughts. To test this, we replace our native teacher (the Qwen2.5-VL vision encoder) with two representative external models:
(1) DINOv2-Large~\cite{oquab2023dinov2} (ViT-L/14), a leading discriminative model; 
(2) SDXL-VAE~\cite{podell2023sdxl}, a generative autoencoder used in diffusion models. As shown in Table~\ref{tab:ablation_teacher}, our endogenous approach consistently outperforms both external variants. While DINOv2 performs well on basic perception tasks, it falls behind on fine-grained benchmarks like VStar. This suggests that aligning latent thoughts with a different feature space creates an \textit{alignment gap}, whereas the native encoder offers naturally aligned supervision. Notably, SDXL-VAE shows a significant performance drop on VisPuzzle, despite good results on MMVP. This highlights a key limitation: VAE latent spaces are optimized for reconstructing pixels and textures, often sacrificing precise spatial logic. Consequently, structural details are lost during reasoning. These results confirm that Endogenous Self-Distillation is not only more efficient but also provides the most robust supervision for visual reasoning.
\begin{table}[htbp]
\centering
\caption{Ablation on teacher selection: \textbf{Endogenous} (Native Encoder) vs. \textbf{Exogenous} (External Experts). While external experts perform adequately on basic perception (MMVP), they struggle with complex spatial reasoning (VisPuzzle), confirming the superiority of endogenous alignment.}
\label{tab:ablation_teacher}
\resizebox{1\textwidth}{!}{%
\begin{tabular}{lcccccc}
\toprule
\textbf{Teacher Model} & \textbf{Type} & \textbf{MMVP} & \textbf{VisPuzzle} & \textbf{VStar} & \textbf{BLINK-J} & \textbf{Avg.} \\
\midrule
DINOv2 (ViT-L/14) & Discriminative & 78.00 & 73.75 & 78.53 & 72.00 & 75.57 \\
SDXL-VAE & Generative & 78.66 & 45.50 & 75.91 & 72.00 & 68.02 \\
\rowcolor[HTML]{E0E0E0}
\textbf{ProLaViT (Native ViT)} & \textbf{Endogenous} & \textbf{79.00} & \textbf{74.25} & \textbf{80.11} & \textbf{76.00} & \textbf{77.34} \\
\bottomrule
\end{tabular}%
}
\end{table}
\begin{table}[htbp]
\centering
\small
\setlength{\tabcolsep}{3pt}
\renewcommand{\arraystretch}{0.9}
\caption{Ablation on reasoning step ordering. Random permutation validates the importance of the canonical coarse-to-fine progression.}
\label{tab:ablation_order}
\resizebox{1\textwidth}{!}{%
\begin{tabular}{lccccc|c}
\toprule
\textbf{Ordering} & \textbf{VisPuzzle} & \textbf{VStar} & \textbf{ChartQA} & \textbf{BLINK-J} & \textbf{CV-Bench} & \textbf{Avg.} \\
\midrule
\rowcolor[HTML]{E0E0E0}
Canonical Order & \textbf{74.25} & \textbf{80.11} & \textbf{78.99} & \textbf{76.00} & \textbf{77.66} & \textbf{77.40} \\
Random Order & 68.00 & 74.50 & 73.50 & 70.66 & 69.14 & 71.16 \\
\quad \textit{Drop} & \textcolor{red!70!black}{-6.25} & \textcolor{red!70!black}{-5.61} & \textcolor{red!70!black}{-5.49} & \textcolor{red!70!black}{-5.34} & \textcolor{red!70!black}{-8.52} & \textcolor{red!70!black}{-6.24} \\
\bottomrule
\end{tabular}%
}
\end{table}

\noindent \boldparagraph{Effect of Reasoning Step Ordering.}
To validate that the causal ordering of reasoning steps is critical, we randomly permute the order of the four steps while keeping all other components fixed. As shown in Tab.~\ref{tab:ablation_order}, the Canonical Order ($z_1 \to z_2 \to z_3 \to z_4$) achieves superior performance (77.40\% average), while Random Order suffers a substantial drop (-6.24\% average), particularly on spatial tasks like VisPuzzle (-6.25\%) and CV-Bench (-8.52\%). These findings provide strong evidence that the \textit{progressive causal structure} is a fundamental requirement for effective visual reasoning.

\noindent \boldparagraph{Efficiency: Latent Reasoning vs.\ Explicit Tool-Use.}
A natural question is whether latent reasoning is preferable to explicitly invoking visual tools at each step. We fine-tune Qwen2.5-VL on identical traces with explicit tool tokens that invoke real operations at inference. As shown in Tab.~\ref{tab:efficiency}, explicit tool-use re-invokes the ViT encoder every step, inducing $3.4\times$ latency and $+117\%$ TFLOPs. In contrast, ProLaViT performs a single ViT pass and reasons entirely in latent space ($1.21\times$ latency, $+12\%$ TFLOPs) while achieving higher accuracy (VStar 80.11 vs.\ 79.60, BLINK-J 76.00 vs.\ 74.00), since latent reasoning avoids the feature loss from iterative image re-encoding. Training requires only ${\sim}$28 GPU-hours on 2$\times$H20, compared to over 120h for generative-CoT methods~\cite{thinkmorph2024}.

\begin{table}[ht]
\centering
\small
\setlength{\tabcolsep}{3pt}
\renewcommand{\arraystretch}{0.9}
\caption{Explicit Tool-Use vs. Latent Distillation.}
\label{tab:efficiency}
\resizebox{0.8\textwidth}{!}{%
\begin{tabular}{lcc|cc}
\toprule
\textbf{Method} & \textbf{VStar} & \textbf{BLINK-J} & \textbf{Latency} & \textbf{TFLOPs} \\
\midrule
Qwen2.5-VL (Base) & 76.44 & 59.33 & 1.00$\times$ & 1.00 \\
Explicit Tool-Use SFT & 79.60 & 74.00 & 3.40$\times$ & 2.17 \\
\rowcolor[HTML]{E0E0E0}
\textbf{ProLaViT (Ours)} & \textbf{80.11} & \textbf{76.00} & \textbf{1.21$\times$} & \textbf{1.12} \\
\bottomrule
\end{tabular}%
}
\end{table}

\section{Conclusion}
In this paper, we presented ProLaViT, a novel framework empowering Multimodal Large Language Models to perform visual reasoning via progressive latent derivation. By transitioning reasoning from discrete text to a structured continuous latent space, ProLaViT effectively bridges the modality gap that limits traditional Chain-of-Thought approaches. Central to our contribution is the Endogenous Self-Distillation mechanism, which enables the model to internalize rigorous algorithmic logic from programmatically synthesized data, thereby eliminating reliance on computationally expensive external vision experts. Furthermore, our proposed Distance-Weighted Diversity Loss mitigates the critical issue of latent collapse, enforcing topological distinctiveness across the reasoning chain to ensure robust, non-redundant visual thoughts. Extensive experiments on perception-intensive benchmarks, including visual search and jigsaw assembly, demonstrate that ProLaViT achieves state-of-the-art performance with superior inference efficiency. While this work focuses on spatial and logical reasoning in static images, the principle of structured latent derivation holds significant potential for broader domains. Future work will extend ProLaViT to temporal video reasoning and complex geometric transformations, paving the way for more generalizable and interpretable multimodal intelligence.


%
%
\bibliographystyle{splncs04}
\bibliography{main}
\clearpage
\setcounter{page}{1}

\begin{center}
    {\Large\bfseries ProLaViT: Learning Progressive Latent Visual\\[0.2em] Thoughts in Structured Latent Space}\\[1em]
    {\large Supplementary Material}\\[0.5em]
\end{center}
\vspace{2em}

\setcounter{section}{0}
\renewcommand{\thesection}{\Alph{section}}

\vspace{-20pt}
    

\section{Implementation Details}

\boldparagraph{Base Model and Parameter-Efficient Tuning.}
We adopt Qwen2.5-VL-7B-Instruct~\cite{qwen25vl} as our base model. To enable efficient adaptation, we apply Low-Rank Adaptation (LoRA) with a rank of 16, a scaling factor $\alpha=32$, and a dropout rate of 0.01. LoRA is applied to all transformer layers within the language model backbone. However, we explicitly exclude the embedding layers and visual branch modules from LoRA updates to preserve their pre-trained representations. During training, both the vision encoder and the language model backbone remain frozen, only the LoRA adapters and a lightweight projection layer are updated.

\boldparagraph{Visual Anchor Configuration.}
In our main experiments, we utilize the native vision encoder of Qwen2.5-VL as the sole source for visual anchors. This encoder produces compact anchor token sequences that serve as visual chain-of-thought checkpoints during inference. To encourage the generation of diverse and non-redundant visual latent states across reasoning steps, we impose a margin-based diversity loss on the anchor token representations with a weight of $\lambda_{\text{div}}=0.2$ and a margin of $\delta=0.8$.

\boldparagraph{Two-Round Training Pipeline.}
We employ a two-round curriculum training strategy:

\begin{itemize}
    \item \textbf{Round 1 (6000 steps).} Starting from the pre-trained Qwen2.5-VL-7B-Instruct, we progressively train the model through three internal stages:
    \begin{itemize}
        \item \textbf{Stage 0 -- VQA Warm-up} (steps 0--1500). Anchor pad tokens are appended to the input to familiarize the model with the anchor token vocabulary, trained via a standard VQA cross-entropy objective.
        \item \textbf{Stage 1 -- Feature Alignment} (steps 1500--4000). The model is trained to explicitly generate the anchor token sequence, supervised by reconstructed visual features from the anchor encoder.
        \item \textbf{Stage 2 -- CoT Training} (steps 4000--6000). The model learns to emit anchor tokens in a \texttt{<think>...</think><answer>...</answer>} format, interleaving visual latent reasoning with the final textual response.
    \end{itemize}
    Upon completion of Round 1, the LoRA weights are merged back into the backbone to produce an intermediate checkpoint.

    \item \textbf{Round 2 (10000 steps).} 
    Building upon the merged Round-1 model, we conduct a joint training phase with restructured stage boundaries (Stage 1: steps 0--3000; Stage 2: steps 3000--6000; VQA-Only: steps 6000--8000; Extended: steps 8000--10000). 
    In the subsequent VQA-Only sub-stage, the model randomly alternates between the anchor-augmented CoT format and the plain VQA format to enhance generalization robustness. Finally, the LoRA weights are merged to yield the production model.
\end{itemize}

\boldparagraph{Optimization Hyperparameters.}
All experiments are conducted on $2 \times$ NVIDIA H20 GPUs utilizing DeepSpeed ZeRO-3 parallelism. We employ the AdamW optimizer with a cosine learning rate schedule, a warm-up ratio of 0.05, and a weight decay of 0.1. The primary learning rate is set to $5\times10^{-5}$, while a reduced rate of $1\times10^{-5}$ is applied to the visual projection layer. The per-device batch size is set to 2, with gradient accumulation steps computed automatically to match the target global batch size. Detailed hyperparameters are summarized in Tab.~\ref{tab:hyperparams}.

\begin{table}[t]
    \centering
    \caption{Hyperparameters for the two-round training pipeline.}
    \label{tab:hyperparams}
    \setlength{\tabcolsep}{10pt} 
    \begin{tabular}{lcc}
    \toprule
    Hyperparameter & Round 1 & Round 2 \\
    \midrule
    Maximum Steps & 6000 & 10000 \\
    Learning Rate & \multicolumn{2}{c}{$5\times10^{-5}$} \\
    Projection LR & \multicolumn{2}{c}{$1\times10^{-5}$} \\
    Weight Decay & \multicolumn{2}{c}{0.1} \\
    Warm-up Ratio & \multicolumn{2}{c}{0.05} \\
    LR Schedule & \multicolumn{2}{c}{Cosine} \\
    LoRA Rank / $\alpha$ & \multicolumn{2}{c}{16 / 32} \\
    LoRA Dropout & \multicolumn{2}{c}{0.01} \\
    Diversity Loss Weight $\lambda_{\text{div}}$ & \multicolumn{2}{c}{0.2} \\
    Diversity Margin $\delta$ & \multicolumn{2}{c}{0.8} \\
    Per-device Batch Size & \multicolumn{2}{c}{2} \\
    Parallelism & \multicolumn{2}{c}{DeepSpeed ZeRO-3} \\
    GPUs & \multicolumn{2}{c}{2 $\times$ NVIDIA H20} \\
    \bottomrule
    \end{tabular}
\end{table}

\section{Training Dataset Construction}

\subsection{Overview.}
To enable the model to internalize algorithmic visual reasoning without
reliance on external tools at inference time, we construct a scalable
programmatic synthesis pipeline that converts single-image VQA samples
into multi-view, multi-operation Chain-of-Thought (CoT) training
instances.
Our training data is drawn from three task domains:
\textbf{Chart Refocus}, \textbf{Jigsaw Assembly}, and
\textbf{Visual Search}.

\subsection{Step 1: Structured CoT Generation.}
Each raw sample consists of a single input image, a natural language
question, and a weakly annotated answer.
To produce operation-grounded reasoning chains, we prompt a large
multimodal model (Gemini) to analyze each sample and output a
structured multi-modal CoT that specifies:
(i) the key visual regions relevant to each reasoning step
    (e.g., chart sub-regions, puzzle pieces, or target objects);
(ii) a sequence of \textbf{candidate visual operations} to be applied
     to those regions; and
(iii) the natural language rationale linking each operation to the
     corresponding reasoning conclusion.

The supported operations include region cropping, bounding-box
annotation, semi-transparent highlighting, spatial grid overlay,
directional arrow and path drawing, and text labeling.
For Jigsaw Assembly, domain-specific operations such as edge
highlighting and counterfactual wrong-assembly illustration are
additionally included.
Operations are further annotated with spatial extents (expressed as
either absolute pixel coordinates or relative image percentages),
color specifications, and sequential dependencies that define the
order in which operations should be applied.

\subsection{Step 2: Programmatic Image Synthesis.}
The structured descriptions from Step~1 are used as input to a
deterministic rendering module that translates textual operation
specifications into concrete image transformations.
The module first parses the operation sequence, resolving spatial
coordinates and dependency relationships to determine whether each
operation should be applied to the original image or to the output
of a preceding operation.
It then executes each operation in order, compositing the results
to produce a chain of intermediate images.
Operations that are independent of one another are applied in
parallel to the original image, while chained operations
(e.g., applying a grid overlay followed by regional highlighting)
are executed sequentially, with each step taking the previous
result as its input.

For each training sample, this process yields a set of
reasoning images---up to four per instance---comprising
the original problem image and one or more intermediate transformed
views.
These images serve as explicit visual checkpoints that anchor each
reasoning step to a concrete perceptual state, forming the
multi-view input consumed by our model during training.

\subsection{Step 3: Metadata Alignment and Quality Filtering.}
After image synthesis, each training instance is assembled with the
following components: the problem image and its associated reasoning
images, the natural language question, the complete multi-step
reasoning chain produced in Step~1, and the final answer.
We apply quality filtering to remove samples in which the structured
CoT generation fails to produce valid output, no executable visual
operation can be identified, or one or more of the synthesized
images is missing or corrupted.
The retained samples from all three task domains are merged into a
unified training set.

\subsection{Task Diversity.}
The three task domains contribute complementary reasoning patterns.
\textbf{Chart Refocus} requires precise quantitative reading and
region-level re-grounding, exercising operations that localize and
annotate specific chart areas.
\textbf{Jigsaw Assembly} demands structural spatial reasoning and
counterfactual analysis of part configurations, relying on operations
that expose assembly errors and directional cues.
\textbf{Visual Search} focuses on target--distractor discrimination
in cluttered scenes, utilizing operations that highlight salient
regions and trace spatial search paths.
Training on this diverse mixture exposes the model to a wide
spectrum of visual operation types and reasoning styles, promoting
generalization of the learned visual chain-of-thought mechanism
across tasks.

\vspace*{4pt} 
\end{document}